\documentclass[runningheads]{llncs}

\usepackage{eccv}

\usepackage{eccvabbrv}
\usepackage{booktabs}
\usepackage{scontents}
\usepackage{graphicx}
\usepackage{array}
\usepackage{tabularx}
\usepackage{algorithm}
\usepackage{algorithmic}
\usepackage{textcomp}
\usepackage{amssymb}
\usepackage{multirow}
\usepackage{xcolor}

\newcommand{\zt}[1]{\textcolor{blue}{[ZT: #1]}}
\newcommand{\ad}[1]{\textcolor{Tan}{AD: #1}}
\newcommand{\as}[1]{\textcolor{red}{AS: #1}}
\newcommand{\kt}[1]{\textcolor{teal}{KT: #1}}
\newcommand{\pv}[1]{\textcolor{magenta}{PV: #1}}
\newcommand{\kv}[1]{\textcolor{orange}{KV: #1}}

\renewcommand{\zt}[1]{}
\renewcommand{\ad}[1]{}
\renewcommand{\as}[1]{}
\renewcommand{\kt}[1]{}
\renewcommand{\pv}[1]{}
\renewcommand{\kv}[1]{}

\usepackage[accsupp]{axessibility}  

\usepackage[pagebackref,breaklinks,colorlinks,citecolor=eccvblue]{hyperref}

\usepackage{orcidlink}

\usepackage{adjustbox}

\begin{document}
\title{DecentNeRFs: Decentralized Neural Radiance Fields from Crowdsourced Images\vspace{-5pt}}

\titlerunning{DecentNeRFs}


\author{Zaid Tasneem\inst{1} \and
Akshat Dave\inst{2} \and Abhishek Singh\inst{2} \and Kushagra Tiwary\inst{2} \and Praneeth Vepakomma\inst{2,}\textsuperscript{3} \and Ashok Veeraraghavan\inst{1} \and Ramesh Raskar \inst{2} \vspace{-6pt}}

\authorrunning{Z. Tasneem et al.}

\institute{Rice University \and
Massachusetts Institute of Technology, \textsuperscript{3} MBZUAI
\\
\email{\{ztasneem,vashok\}@rice.edu}, \email{\{ad74,abhi24,vepakom,ktiwary,raskar\}@mit.edu}}

\maketitle

\begin{figure}
\centering
\includegraphics[width=\textwidth]{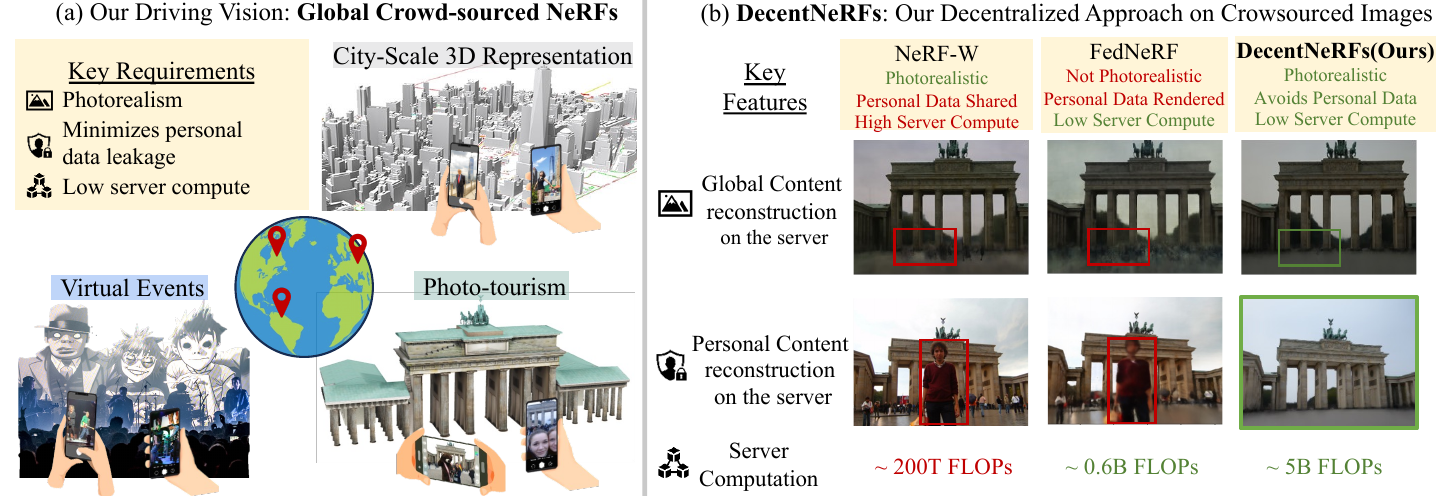}
\captionof{figure}{\textbf{Overview of our approach:} Our framework leverages geotagged images of locations siloed worldwide in user photo galleries. It constructs neural radiance fields (NeRFs) for immersive 3D viewing in a decentralized manner, minimizing the reconstruction of personal content at the server. (\textbf{Left}): We showcase the potential to generate global-scale 3D scene representations for applications like city-scale NeRFs, immersive past event experiences, and virtual photo-tourism. We highlight the capture of diverse viewpoints across users. (\textbf{Right}): We compare our approach against centralized and existing decentralized baselines for a photo-tourism example. We demonstrate our approach's optimal tradeoffs between low-server compute and photorealism while minimizing personal content reconstruction on the server. 
}
\label{fig:teaser}
\end{figure}
\label{fig:teaser}

\begin{abstract}
\vspace{-2em}
Neural radiance fields (NeRFs) show potential for transforming images captured worldwide into immersive 3D visual experiences. However, most of this captured visual data remains siloed in our camera rolls as these images contain personal details. Even if made public, the problem of learning 3D representations of billions of scenes captured daily in a centralized manner is computationally intractable. Our approach, DecentNeRF, is the first attempt at decentralized, crowd-sourced NeRFs that require $\sim 10^4\times$ less server computing for a scene than a centralized approach. Instead of sending the raw data, our approach requires users to send a 3D representation, distributing the high computation cost of training centralized NeRFs between the users. It learns photorealistic scene representations by decomposing users' 3D views into personal and global NeRFs and a novel optimally weighted aggregation of only the latter. We validate the advantage of our approach to learn NeRFs with photorealism and minimal server computation cost on structured synthetic and real-world photo tourism datasets. We further analyze how secure aggregation of global NeRFs in DecentNeRF minimizes the undesired reconstruction of personal content by the server.


\end{abstract}    



\section{Introduction}
\label{sec:intro}


Every day, more than 5 billion photos are captured worldwide, comprising multiple viewpoints of every monument, skyscraper, cafe, and concert on Earth. Neural radiance fields (NeRFs) present an exciting opportunity to process this massive data into immersive visual experiences at a global scale. However, most of these images remain siloed in personal camera rolls. Less than 2\% of these captured photos are ever posted on the internet \cite{photutorial2023}. Even if these personal images were made public, learning NeRFs for billions of scenes captured daily at a global scale in a centralized fashion is computationally intractable. Therefore, to build immersive visual experiences at a global scale, NeRFs must be decentralized to handle high compute needs and avoid the undesired reconstruction of personal content by a central entity, all the while ensuring photorealism (Fig.~\ref{fig:teaser}(a)).

Our framework, DecentNeRF, is the first attempt towards decentralized NeRFs from crowd-sourced images. Images in public spaces are often composed of \textit{global} content,e.g., a monument that we would like to share with the world and a \textit{personal} content, such as a friend posing in front of the monument that we would like to keep to ourselves. Our key insight is that often global content is \textit{static} across users, and the personal content is \textit{dynamic}, i.e., varies from user to user. This association of global as static and personal as dynamic allows our approach to perform global-personal separation in the captured images. Our approach enables sharing only the global scene-specific 3D representations across users instead of sharing the combined (global+dynamic) image as in a conventional NeRF pipeline. In doing so, the server avoids the cost of centrally training NeRFs by distributing the NeRF training computation across users. It also minimizes the reconstruction of the undesired occlusions of personal user-specific content at the server.

For a particular scene, we model the multi-view visual data as a combination of a \textit{global radiance field} for the 3D scene of interest and a \textit{personal radiance field} for the user's personal information (transient across users). We propose federated learning \cite{kairouz2021advances} procedure to learn the global radiance field across users by aggregating only the user's global radiance field model (locally trained). Instead of uniformly averaging the users' weights as typical in federated learning \cite{konevcny2016federated}, we propose a novel federation procedure where the per-user scaling is learned implicitly to maximize visual fidelity. To prevent the server from accessing the individual user's global radiance fields, we use a secure multi-party computation (SMPC) protocol  \cite{burkhart2010sepia} for aggregation. The secure aggregation minimizes the reconstruction of personal content by the server compared to existing approaches during initial rounds of federation.


Our proposed framework, DecentNeRF is the first work to analyze the decentralization aspects of radiance fields for real-world crowd-sourced images.  NeRF-W \cite{martin2021nerf} achieves the high visual quality of the public global scene from in-the-wild crowd-sourced images but requires all the personal user images to be transferred to a central server for training. This results in personal content directly being accessed by the server and high server compute(Fig.~\ref{fig:teaser}(b)). We address both these limitations in our work. Works on federated learning of NeRFs \cite{holden2023federated,suzuki2023federated} demonstrate model compression and large scene modeling capabilities but assume the images of static scenes devoid of users' personal information. When dealing with crowd-sourced data, these approaches not only allow the server to access personal information without a secure aggregation protocol but also result in inaccurate reconstructions (Fig.~\ref{fig:teaser}(b)). Our approach demonstrates high visual quality in the global content compared to existing decentralized approaches, with $\sim10^4\times$ lower compute than the centralized approach(Fig.~\ref{fig:teaser}(b)).



To summarize, we make the following contributions.

\begin{itemize}
    \item We introduce DecentNeRF, the first approach to address the challenges of learning global 3D scene representations at scale from crowd-sourced images in a decentralized manner.
    \item Our method uses public-global separation and a novel learned federation scheme to achieve high-quality reconstruction of 3D scenes with very low server computing compared to prior works (Fig.~1(b)).  
    \item  Our method outperforms existing federated learning NeRF techniques on scene reconstruction quality while requiring $\sim10^4\times$ lower compute than centralized approaches. We demonstrate this in simulated (Table \ref{tab:blender_baseline}) and real-world crowdsourced scenes (Table \ref{tab:photo_qual}). 
    \item We provide an analysis on how secure aggregation of user's global NeRFs in DecentNeRF reduces the reconstruction of personal content on the server Fig. \ref{fig7:blender_privacy} for both datasets compared to prior works.
\end{itemize}




\section{Related Work}
\label{sec:relatedWork}

\begin{figure}[t!]
    \centering
    \includegraphics[width=\textwidth]{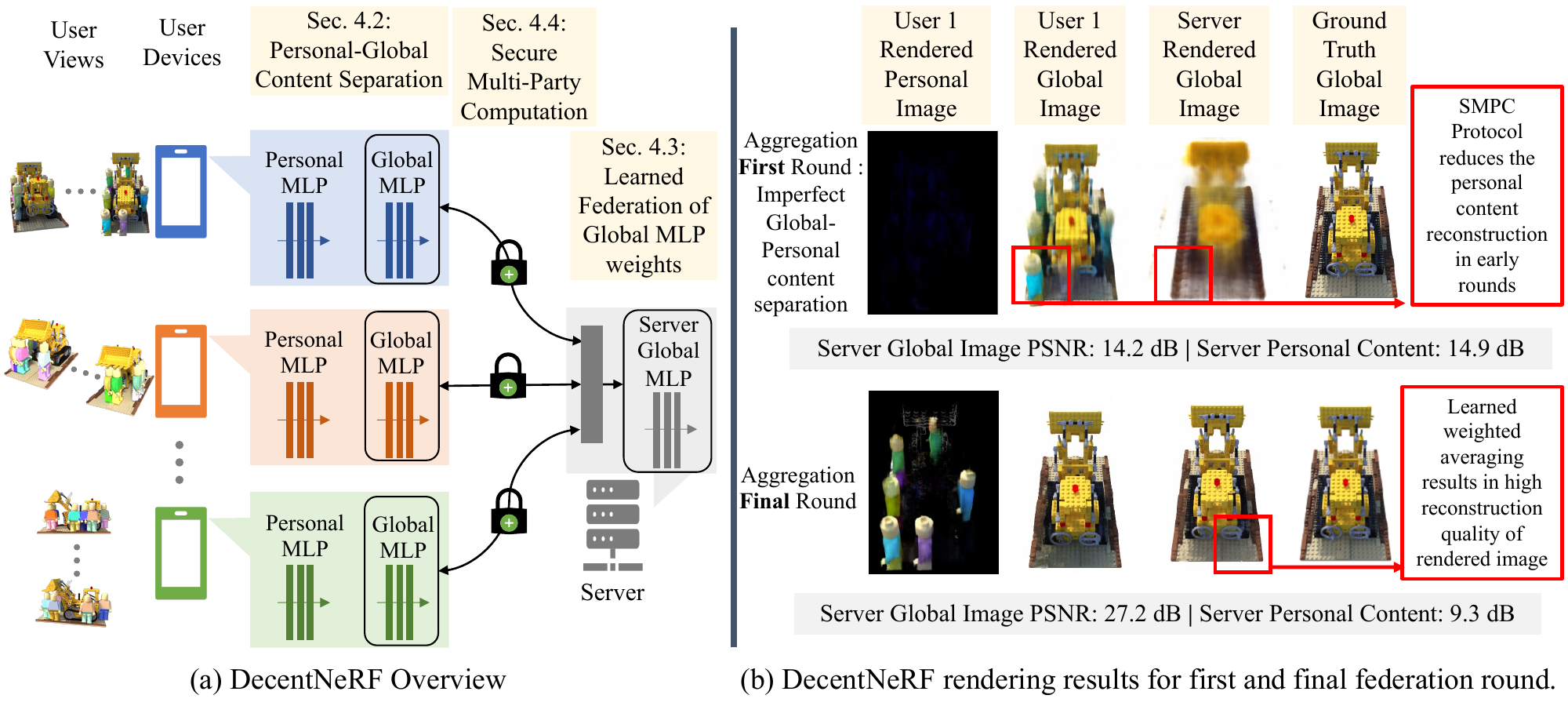}
    \caption{\textbf{Key features of our approach} (a) Overview: Personal and global MLPs are trained on user devices to separate personal and global content from local images. After each training round, the server performs a learned federation of users' global MLPs using a secure MPC protocol and distributes the updated global MLP back to each user. (b) Results on reducing personal content leakage: We notice that users' global MLPs contain personal content during the initial rounds. Our secure MPC protocol ensures the server only sees the averaged global MLP from which the rendering of users' personal content is minimal. Over federation rounds, global and personal MLPs separate content through learned weighted averaging, enabling high-fidelity rendering from the server's global MLP.} 
    \label{fig:key_insights}
\end{figure}

\noindent\textbf{3D from Unstructured Image Collection.} Generating 3D scene representations for novel-view synthesis from an extensive collection of unstructured images \cite{phototourism} has been explored in \cite{martin2021nerf, chen2022hallucinated, yang2023cross,li2020crowdsampling}. However, these prior works are centralized by nature - all the captured images are sent to the server for training NeRFs. Our work, DecentNeRF, diverges from traditional work by focusing on decentralization to 1) distribute the training compute to the users and thus scale to billions of scenes and 2) avoid aggregating users' images, which could contain personal details. Our vision assumes NeRFs can be trained without users' attrition of mobile devices. Recent advancements in mobile NeRF renderings, such as \cite{chen2022mobilenerf,cao2023real}, demonstrate the promise for eventual mobile deployment.


\noindent \textbf{Federated Learning.} In Federated learning (FL)\cite{konevcny2016federated,kairouz2021advances}, each client device trains the model parameters on-device using its own local data. The server then performs a weighted average of the models to obtain a global model and this process continues till convergence. The works in  \cite{holden2023federated,suzuki2023federated} apply federated learning in the context of NeRFs for achieving the utility of 3D scene modeling. Since, NeRFs are 3D representations sharing them instead of raw data would still lead to reconstruction of the personal data. We instead model the local 3D scene representation as a combination of a global radiance field for the 3D scene-specific detaisls and a personal radiance field for the user-specific information.  This decoupling approach is reminiscent of parameter decoupling in personalized federated learning~\cite{tan2022towards}. However, unlike personalized federated learning, we consider each sample to be composed of personal and non-personal information. Hence, the goal is still to learn a single global consistent scene across multiple users. 
\newline \noindent \textbf{Secure Aggregation.} Accessing individual model updates from the users in FL could lead to data reconstruction attacks \cite{yin2021see, jeon2021gradient}. Therefore, secure aggregation has been proposed to average model updates such that only the final averaged weights are revealed to the untrusted central server. This is accomplished by encrypting individual model updates by each user such that only the final average can be decrypted. To compute the average of encrypted user models, existing techniques rely on primitives such as secure multi-party computation(SMPC). It has been studied extensively in FL since the first practical construction by~\cite{bonawitz2017practical}. Recent works have explored different dimensions of secure aggregation such as communication~\cite{choi2020communication} and
 synchronization~\cite{so2022lightsecagg,nguyen2022federated}. We use SMPC-based secure aggregation \cite{lindell2020secure} as a building block to perform weighted averaging over encrypted model updates.
\section{Key Insights}
\label{sec:key_insights}

\zt{add a paragraph and images that say what is personal and what's global for the three datasets}
Consider a scenario where users visit a famous restaurant in town at different times over months capturing images that are now saved on their personal photo galleries. In addition to visual content depicting the restaurant scene, these images likely contain personal content that users would not want to share publicly. We aim to develop a decentralized approach where a server can learn a 3D representation of the restaurant, given such a cluster of users and their captured images. This is a challenging problem as the learned scene representation must encode both the appearance and 3D structure of the scene while not revealing the personal image content to the server. To create a global-level 3D representation, we can repeat this process for millions of users and locations (restaurants, monuments, etc.) which puts a compute constraint on the server learning the scene representation. This section highlights the key insights enabling our approach to achieve photorealistic 3D reconstruction with minimal server computing and undesired reconstruction of personal content.

\noindent \textbf{Decentralization through Federated 3D Scene Representations.} Neural radiance fields (NeRFs) excel at encoding 3D scene information using multilayer perceptrons (MLPs) \cite{nerf}. Existing decentralized solutions collaboratively learn a shared NeRF MLP representation with each user's local views \cite{holden2023federated}. The users can refine the MLPs locally and in parallel, offloading compute from the server. The server only needs to aggregate user MLP updates into a combined shared MLP and transmit this back to users for further refinement. Over multiple federation rounds, this approach aims to reconstruct a 3D scene with the shared MLP. Such a federated method requires orders of magnitude less server computation than centralized approaches, aligning with our decentralization goal. 

\noindent \textbf{Existing Federated NeRF performs poorly on crowdsourced images.} However, existing federated NeRF approaches assume input view consistency - that any 3D point observed from users' images is static. The underlying assumption is that all users took all images at the same instant, only capturing the global scene content and avoiding personal data. These assumptions do not hold for crowdsourced images taken over months and contain personal content like users, their food, or credit cards which are transient across users. Violations of these assumptions would hamper reconstruction quality (Fig. \ref{fig:blender_qualitative_utility}) and leak personal content from the shared MLPs (Fig. \ref{fig7:blender_privacy}). We now provide key insights into how we can exploit the structure of these violations to learn photorealistic global 3D scene content in a decentralized manner.  


\begin{enumerate}
    \item \textbf{Personal-Global Content Separation.} Only the global scene-specific content is 3D view-consistent (static) across users such as the columns and most of the restaurant's interior. By definition, all other 3D content would be transient across users, be it non-personal like the wait staff or personal and sensitive like the user, their, or the credit card on the table. To leverage the juxtaposition between scene-specific and user-specific content, we propose encoding 3D appearance between personal and global MLPs in Sec. \ref{sec:client_scene_sep}, capturing personal and global content, respectively. Only the global MLP is federated at the server to form the combined global MLP. This allows for high-quality reconstruction of global content over multiple rounds of federation as shown in Fig. \ref{fig:key_insights}b (bottom). 
    \item \textbf{Learned Federation.} Users likely have different data distributions - number of views, disparity, and user/scene content ratios. Naive federated averaging of global MLP \cite{konevcny2016federated} is suboptimal as shown in Sec. \ref{sec:results} and Fig. \ref{fig6:smart_aggregation}. We instead learn aggregation weights over federation rounds in Sec. \ref{sec:learned_wts} for optimal reconstruction quality. 
    \item \textbf{Secure Multi-Party Computation (SMPC).} We notice that during the initial rounds, the user global MLPs encode both personal and global content as highlighted in Fig. \ref{fig:key_insights}b (top). This is because users initially have no notion of global or personal content without federation across users. If the server has access direct access to the users' global MLPs, which is the case for existing FL NeRF methods \cite{holden2023federated}, it can faithfully render the personal content as shown in user 1 rendered global image in Fig. \ref{fig:key_insights}(b). To prevent the server from outright accessing the individual user's global MLPs, we use SMPC aggregation described in Sec. \ref{sec:mpc}. Now, the server can only access averaged global MLP at each round. We analyze how these securely aggregated server global MLPs have minimal reconstruction of personal content during initial rounds in Fig. \ref{fig:key_insights}(b) (Top row-third column). We further analyze the reconstruction of personal content from aggregated server global MLPs for the case of varying numbers of users and overlap of users' views in Section \ref{sec:num_users_recons_personal_content}. \vspace{-5pt}
\end{enumerate}

\noindent \textbf{Overall Compute.}  It is important to note that decentralized NeRF approaches, including ours, would have total computing (server+user) similar to a centralized approach (only server) for training NeRFs. However, in the case of decentralized systems, the training of NeRFs is distributed to the users to reduce server computing significantly. Our approach's compute and communication overheads are discussed in detail in the supplement.
\begin{figure}[t!]
    \centering
    \includegraphics[width=0.9\linewidth]{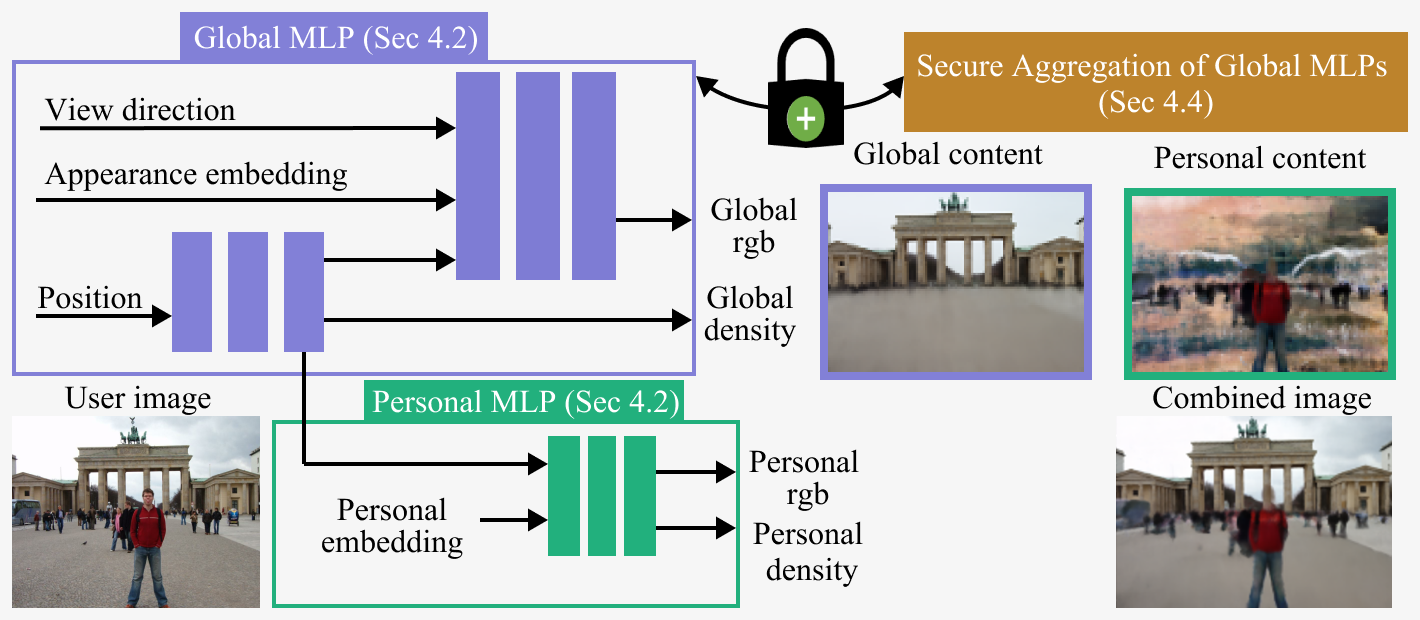}
    \caption{\textbf{DecentNeRF architecture:} On user devices, we consider NeRFs with the following architecture where the personal MLP is always local to the user and the weights of global MLP are securely aggregated at the server. We also highlight what DecentNeRF learns to represent in its Global and Personal MLPs. 
    }
    \label{fig:method}
\end{figure}
\section{Our Approach}

In this section, we discuss the assumptions our approach makes, develop each of the three key insights described in Sec. \ref{sec:key_insights}, and show how they come together in our algorithm.

\subsection{Preliminaries}
\label{sec:prelim}
We state the assumptions necessary for our approach:

\noindent\textbf{1. Personal content is transient across users.} We assume not all clients have taken images at the same instant, which is a realistic assumption for crowdsourced images. It ensures that the personal content is transient and not static across users like the global content.
\noindent\textbf{2. Sufficient overlap in views.} Our approach assumes that no single user has views that don't overlap with another user. We expand more on this in Sec. \ref{sec:num_users_recons_personal_content}.
\noindent\textbf{3. Known camera poses.} Our NeRF pipeline assumes known, accurate camera poses.  In centralized settings, structure from motion (SfM) on image feature descriptors like SIFT can derive relative camera poses using COMLAP \cite{schonberger2016structure}. Decentralized SfM is an active research area \cite{dusmanu2021privacy,asadi2024di} and beyond the scope of this work.

\noindent\textbf{Threat model.} From a user's perspective, the server and other users are untrusted. However, we assume no collusion between users to leak more information than possible individually. Unlike differential privacy~\cite{dwork2006calibrating}, which prevents identity leakage, we focus on preventing the reconstruction of user-specific semantic personal information by the server. The server can try to reconstruct a target user's personal information from the averaged model updates. We currently limit our scope to honest-but-curious servers who follow protocols but attempt to reconstruct raw data based on observations, leaving other attack types for future work.

\noindent \textbf{NeRF Backbone \cite{nerf}} At its core, NeRF trains a multi-layer perceptron (MLP) that takes a 5D input vector: a 3D spatial coordinate $\mathbf{x} = (x, y, z)$ within the scene, and pitch $\theta$ and yaw $\phi$ parameters of the viewing direction $\mathbf{d}$. The MLP outputs a scalar density $\sigma$ and an RGB color vector $\mathbf{c} = (r, g, b)$. NeRF rendering synthesizes images via volume rendering, sampling predicted color and density at many points along rays traversing the scene. For a particular ray corresponding to a pixel, the MLP predicts a color vector $\mathbf{c} = (r, g, b)$ and compares it with the ground truth RGB value, using the loss to train the MLP weights. We refer the reader to \cite{nerf} for more details. 

\subsection{Personal-Global Content Separation}
\label{sec:client_scene_sep}
Consider a scene with $K$ known users. For each user $k$, we model the scene using global and personal MLPs (Fig.~\ref{fig:method}) with the MLP weights denoted as $\mathbf{g}_k$ and $\mathbf{p}_k$ respectively. The personal MLPs are kept native to the user's device. In contrast, the global MLP captures shared global content (such as the landmark) across users via federation. Each MLP outputs its scalar density $\sigma$ and an RGB color $\mathbf{c} = (r, g, b)$, so the rendered image is an alpha-composite of both output images.

Our architecture is inspired by NeRF in the Wild (NeRF-W) works in \cite{martin2021nerf,chen2022hallucinated}, with each user essentially running NeRF-W locally. The global and personal MLPs are analogous to the static and transient MLPs in the context of a single user. However, the global-personal definition better suits the decentralized setting since NeRF-W's transient MLP will not separate all of the personal content from the user. Imagine a subject being recorded in front of a monument with multiple images. Since the subject would be static across all views within the user's images, it will not appear in the transient MLP even though it is transient across users. Our federation of global MLP across users over multiple rounds can separate this user-static but transient across user-specific content into the personal MLP branch (Fig. \ref{fig:method} (inset)) and lead to high-quality reconstructions

\zt{this can be addressed in prior works or deleted completely}
\noindent Prior works removing content from NeRFs use segmenting and inpainting personal content (e.g. people) \cite{weder2023removing}. In our case, the personal-global MLP separation helps us implicitly separate the personal and global content.

\begin{figure}[t!]
    \centering
    \includegraphics[width=\linewidth]{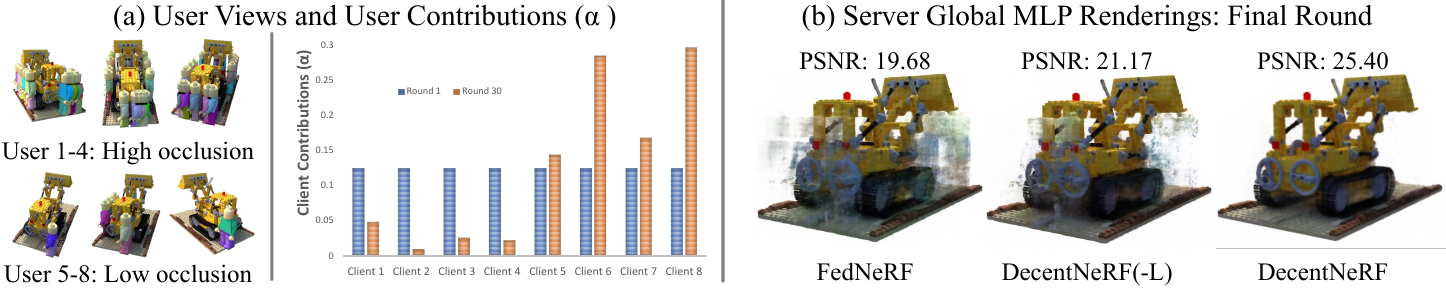}
    \caption{\textbf{Ablation on Learned Federation}: We demonstrate that with Learned Federation of Global MLPs, our approach (DecentNeRF) learns to weigh clients with less occlusion over 30 rounds of training which leads to better reconstruction quality overall compared to one with FedAvg aggregation scheme i.e. DecentNeRF(-L) and FedNeRF.}
    \label{fig6:smart_aggregation}
\end{figure}

\subsection{Learned Federation of Global MLPs}

\label{sec:learned_wts}
\zt{needs more clarity}
The global MLP weights at any aggregation round $m$, are defined as $\mathbf{g}^{(m)} = \sum{\alpha_k^m \cdot {\mathbf{g}}_k^m}$. In a naive federated learning approach, all users are weighted equally, $\alpha_k= \frac{1}{K}$. FedAvg \cite{konevcny2016federated} defines the users based on the number of pixels in their image data, $\alpha_k = \frac{p_k}{\sum_{k=1}^{K} p_k}$, where $p_k$ is the total number of image pixels in the user data.   

These existing definitions lead to suboptimal reconstruction performance as shown in Fig.~\ref{fig6:smart_aggregation} because all users have different data distributions -  number of views, view disparity, user/scene content ratios and would be able to do different levels of global-personal content decomposition to benefit the server's global representation. We aim to learn the $\alpha$ implicitly over different merge rounds by the following equation 

\vspace{-1em}
\begin{equation}
\label{eq:iterative_update}
    \alpha_k^{(m)} = \alpha_k^{(m-1)} - \eta \frac{\partial L}{\partial \alpha_k^{(m-1)}}
\vspace{-0.2em}
\end{equation}
where \(\eta\) is the weighted averaging learning rate, and the cumulative loss \(L\) is defined as the sum of losses for each user \(k\), \(L = \sum_k L_k\), where \(L_k(g_k^{(m-1)}, p_k^{(m-1)}, I_{\text{GT}})\) is the training loss defined in \cite{martin2021nerf}.

The loss implicitly rewards the user's weightage, $\alpha_k$ when they do a better global-personal decomposition in the previous round's aggregation. Combining the above with the update equation for \(\alpha_k\), 
\begin{equation}
\vspace{-0.5em}
    \frac{\partial L}{\partial \alpha_k}^{(m-1)} = \left( \sum \frac{\partial L_k}{\partial \mathbf{g}^{(m-1)}} \right) \cdot \mathbf{g}_k^{m-1}
\vspace{-0.2em}
\end{equation}

Beyond the first round of aggregation, each user sends its own \(\frac{\partial L_k}{\partial \mathbf{g}^{(m-1)}}\), which is aggregated by the server and sent back to the users. This aggregated value gets multiplied with the previous round's user weights at the user to update each user's \(\alpha_k\) at every round.



\begin{figure*}[t]
    \centering
    \includegraphics[width=\textwidth]{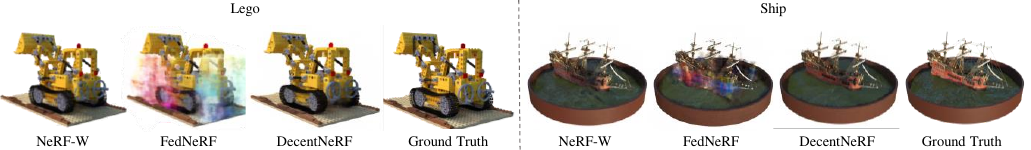}
    \caption{\textbf{Qualitative results on Novel Blender dataset}: DecentNeRF simultaneously removes unwanted personal content (`lego persons') while reconstructing fine details of global content (`lego excavator' and `ship'). In contrast, FedNeRF\cite{holden2023federated} hallucinates `blobs' where personal content exists in user views.}\label{fig:blender_qualitative_utility}
\end{figure*}

\begin{table*}[ht]
\scalebox{0.8}{{
\begin{tabularx}{\textwidth}{l|ccc|c|ccc|c|}
 & \multicolumn{4}{c|}{Lego} & \multicolumn{4}{c|}{Ship}\\
\cline{1-9}
 & \multicolumn{3}{c|}{Photorealism} & Decentralization  & \multicolumn{3}{c|}{Photorealism} & Decentralization\\
  & PSNR$\uparrow$ & SSIM$\uparrow$ & LPIPS$\downarrow$ & Server FLOPs$\downarrow$ & PSNR$\uparrow$ & SSIM$\uparrow$ & LPIPS$\downarrow$ & Server FLOPs$\downarrow$\\
NeRF-W & \textcolor{Blue}{27.09} & \textcolor{Blue}{0.934} & \textcolor{Blue}{0.0513} & $\approx$ 130 T & \textcolor{Blue}{26.35} & \textcolor{Blue}{0.8661} & \textcolor{Blue}{0.1765} & $\approx$ 220T\\
FedNeRF & 17.7 & 0.858 & 0.2003 & $\approx$ \textcolor{Blue}{0.6} B & 20.92 & 0.8281 & 0.2368 & $\approx$ \textcolor{Blue}{1 B}\\
DecentNeRF &	\textcolor{BurntOrange}{26.25} &	\textcolor{BurntOrange}{0.92}	& \textcolor{BurntOrange}{0.0652} & $\approx$ \textcolor{BurntOrange}{4.8 B}	& \textcolor{BurntOrange}{24.58} &	\textcolor{BurntOrange}{0.8554}	& \textcolor{BurntOrange}{0.1901}	& $\approx$ \textcolor{BurntOrange}{8 B}
\end{tabularx}}}
\caption{\textbf{Quantitative results on Novel Blender dataset}:
Best and second best results for  are highlighted in blue and orange respectively. A centralized approach \cite{martin2021nerf} achieves the best photorealism performance while using ~\textbf{$10^5$} more server compute than decentralized approaches. FedNerf \cite{holden2023federated} even though it only is 8x server compute cheaper than DecentNeRF and performs poorly on photorealism metrics, In contrast, DecentNeRF obtains comparable photorealism to the centralized approach with several orders of magnitude lower sever compute.}
\label{tab:blender_baseline}
\end{table*}

\subsection{Secure Multi-Party Computation (SMPC)}
\label{sec:mpc}

From the user's perspective, in every round, there are exactly two kinds of messages that each user \(k\) communicates with the server - 1) weights of the global MLP, \(\mathbf{g}_k\), and 2) gradients with respect to the per-user loss, \(\frac{\partial L_k}{\partial \mathbf{g}^{(m-1)}_k}\). They also compute 3) gradient updates of \(\alpha\) \(\left(\frac{\partial L^{(m-1)}}{\partial \alpha_k}\right)\). We aim to obfuscate the user communications such that only the final result aggregated over all the users is visible to the server. From the server's perspective, it needs to compute - 4) the weighted average of the obfuscated weights sent by the users \(\mathbf{g}^{(m)}\), and the cumulative loss \(\sum \frac{\partial L_k}{\partial \mathbf{g}^{(m-1)}}\). 
Each of these four functional requirements is satisfied using secure multi-party computation \cite{lindell2020secure}. 
We now describe the protocols in more detail.

$\mathsf{send\_wts(\mathbf{g}^{(m)}_k)}$: Each user obfuscates its local weights (multiplies with $\alpha_k$) by applying a random mask or encrypting data using additively homomorphic encryption techniques. This randomization ensures that when the obfuscated weights are aggregated, the original values can not be recovered from those weights while providing the means to calculate the weighted average accurately.

$\mathsf{weight\_avg(\mathbf{g})}$: This function executes the computation of the weighted average of the obfuscated weights sent by the users ($\mathbf{g^{(m)}}$). By employing secure multi-party computation techniques, the server aggregates the masked weights received from all users. 

$\mathsf{send\_grads()}$ This function is similar to $\mathsf{send\_wts}$ in its design as each user communicates its local gradients by performing randomized obfuscation to its gradients in a similar way it performs obfuscation of the weights.

$\mathsf{compute\_grads()}$ This function is relatively simple where each user calculates the update step described in equation (2) and then updates $\alpha_k$ accordingly.

\subsection{Summarizing our algorithm}
Putting together the building blocks developed above, our algorithm can be summaries as (We refer the reader to the supplementary material for pseudo-code of the detailed algorithm):
\begin{itemize}
    \item Assumes random initialization of global \(\mathbf{G^{(0)}}\) and personal MLP weights \(\mathbf{P^{(0)}}\) at all \(K\) users.
    \item (On the user) For each round the users return the current \(\mathbf{g}_k^{m}\) and previous rounds global MLP along with loss gradients of previous round global MLP weights \(\frac{\partial L_k}{\partial \mathbf{g}^{(m-1)}}\). They also calculate the gradient updates for $\alpha_k$.
    \item (On the server) For each round the weights and previous rounds gradients are securely aggregated and sent back to the user.
\end{itemize} 
\vspace{-5pt}

\section{Experimental details}

\subsection{Datasets}
\label{sec:datasets}

\textbf{Novel Blender Dataset.} To do a controlled study of the effects of varying occluding personal content, number of users, and overlap of user views, we introduce our own synthetic dataset. We do it by adding personal content in the form of Lego people to the original blender dataset from \cite{nerf}. The equivalent of this would be actual people/occlusions in a real-world setting. Unless mentioned otherwise, the dataset comprises 360-degree rotation with non-IID partitioning of 18 degrees in yaw angle rotation of 100 images distributed equally among 20 users. Each view contains six Lego people, with one static person per user's view representing personal content. More details and examples from the novel dataset are shown in the supplementary. 


\noindent \textbf{Phototourism Dataset.} To evaluate the decentralization benefits of our approach on a real-world dataset we also evaluate 3 scenes on novel-view synthesis from unconstrained image collections: Brandenburg Gate, Trevi Fountain, and Sacre Couer\cite{phototourism}. Our setup has 20 clients each with 10 images of 2x downsampled resolution. Please refer to the supplementary for more details. 


\subsection{Baselines}
We compare our solution with a centralized approach (NeRF-W\cite{martin2021nerf}) and an existing decentralized approach (FedNeRF\cite{holden2023federated}). We also run an ablation of our approach without the learned federation federation.
\subsection{Evaluation Metrics}

\noindent \textbf{Photorealism.} We use PSNR, SSIM, and LPIPS on the global MLP renders as described in prior works \cite{nerf,holden2023federated} for our novel blender dataset. For Phototourism, we use the same protocol as in \cite{martin2021nerf} for fitting the appearance of the left half of the validation images and evaluating on the right half for DecentNeRF and NeRFW.
\begin{figure*}[t]
    \centering
    \includegraphics[width=\textwidth]{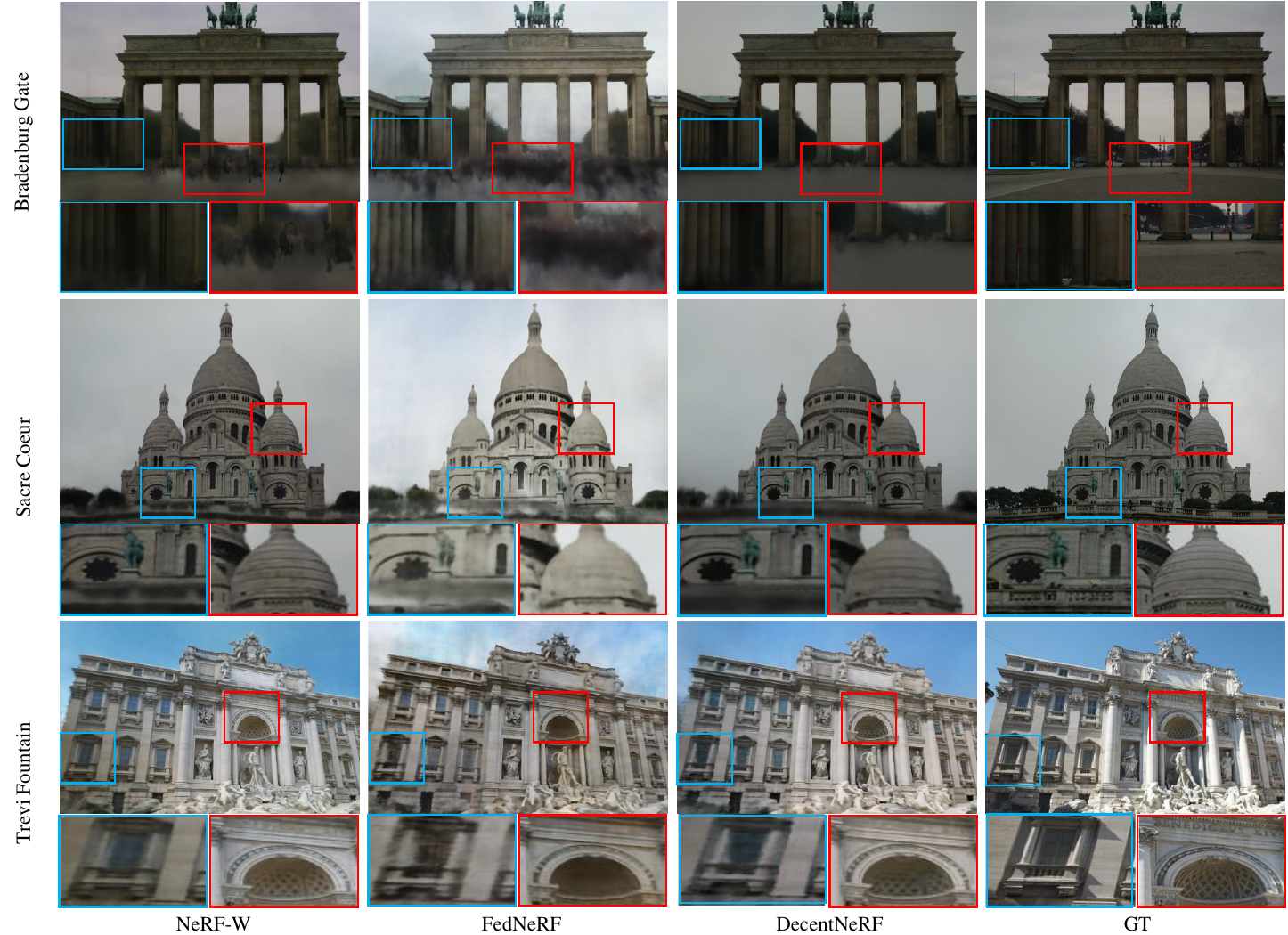}
    \caption{\textbf{Qualitative results on Phototourism\cite{phototourism} dataset}: DecentNeRF is simultaneously able to reduce personal content (top row) from the scenes and capture fine details (middle-bottom). It does well in reconstruction in areas where personal content is present in user images highlighted by blue squares specially in top and middle rows.}
    \label{fig:photo_qualitative}
\end{figure*}

\noindent \textbf{Decentralization.} To do a fair comparison we use the FLOPs computation from the SOTA NeRF training \cite{muller2022instant} and calculate server and per-user compute. Server compute includes aggregation, reconstruction cost for SMPC protocols, and user weight update. For users, it is NeRF training compute. For our application only server compute is relevant and analyzed in the main paper. Please refer to the supplementary material for more details on server and client costs including communication costs
     


\begin{table*}[ht!]
\scalebox{0.65}{
\begin{tabularx}{\textwidth}{l|ccc|c|ccc|c|ccc|c|}
 & \multicolumn{4}{c|}{Bradenburg Gate} & \multicolumn{4}{c|}{Trevi Fountain} & \multicolumn{4}{c|}{Sacre Coeur}\\\cline{1-13}
 & \multicolumn{3}{c|}{Photorealism} & Decentralization  & \multicolumn{3}{c|}{Photorealism} & Decentralization & \multicolumn{3}{c|}{Photorealism} & Decentralization \\
  & PSNR$\uparrow$ & SSIM$\uparrow$ & LPIPS$\downarrow$ & Server FLOPs$\downarrow$ &  PSNR$\uparrow$ & SSIM$\uparrow$ & LPIPS$\downarrow$ & Server FLOPs$\downarrow$ & PSNR$\uparrow$ & SSIM$\uparrow$ & LPIPS$\downarrow$ & Server FLOPs$\downarrow$\\
  NeRF-W  & \textcolor{blue}{25.18}	& \textcolor{blue}{0.901} &	\textcolor{blue}{0.1927} & $\approx$ 200 T & 	\textcolor{blue}{21.49}	& \textcolor{blue}{0.7206} &	\textcolor{blue}{0.2866} & $\approx$ 250 T &	\textcolor{blue}{20.58}	& \textcolor{blue}{0.7835}	& \textcolor{blue}{0.2305} & $\approx$ 280 T \\
  FedNeRF & 18.92 &	0.8158 &	0.3528 & $\approx$ \textcolor{Blue}{0.6 B} &	16.55 &	0.5774	& 0.4352 & $\approx$ \textcolor{Blue}{1.2 B} &	15.41 &	0.6555	& 0.4072 & $\approx$ \textcolor{Blue}{1.5 B}\\
  DecentNeRF &	\textcolor{BurntOrange}{24.62} & \textcolor{BurntOrange}{0.8802} & \textcolor{BurntOrange}{0.2571} & $\approx$ \textcolor{BurntOrange}{5 B} & \textcolor{BurntOrange}{20.61} & \textcolor{BurntOrange}{0.6501} & \textcolor{BurntOrange}{0.3963	} &   $\approx$ \textcolor{BurntOrange}{9 B} & \textcolor{BurntOrange}{19.22} & \textcolor{BurntOrange}{0.7259} & \textcolor{BurntOrange}{0.3126} & $\approx$ \textcolor{BurntOrange}{10 B}
\end{tabularx}
}
\caption{\textbf{Quantitative performance on Phototoursim dataset\cite{snavely2006photo}}: Best and second best results are highlighted in blue and orange respectively. DecentNeRF is effective for real-world crowd-sourced datasets compared to other decentralized approaches \cite{holden2023federated} while using $10^{4}$ less compute than centralized approaches \cite{martin2021nerf}. 
\vspace{-10pt}
}
\label{tab:photo_qual}
\end{table*}

\noindent \textbf{Personal content.} As we know the ground-truth mask of the personal content, we use NeRF-centric reconstruction metrics like PSNR to measure personal content for the novel blender dataset. These empirical evaluations have been used in other scene inversion works in \cite{pittaluga2019revealing,ng2022ninjadesc,tasneem2022learning}. For the real-world phototourism dataset, we go one step further and use a pre-trained Faster R-CNN ResNet-50 FPN\cite{girshick2015fast} object detector and report the number of people detected as a metric of personal content. We render views for the server-accessible MLPs which would be users' Global MLPs for FedNeRF and Server Global MLP for DecentNeRF.

\subsection{Implementation Details}
We use a PyTorch Lightning implementation of NeRF-W \cite{nerf_pl_nerfw} for centralized evaluation. We implement the secure aggregation protocol, SecAgg \cite{bonawitz2017practical}, using Flower\cite{beutel2020flower}. Our approach differs from FedNeRF \cite{holden2023federated} as we employ a Personal MLP, appearance embedding, secure aggregation, and learned weighted averaging. Please refer to our supplementary material
for training and model details. \zt{add labels}



\section{Results and Analysis}
\label{sec:results}
\subsection{Ablation on Learned Federation}
To demonstrate the advantage of Learned Federation of weights in terms of photorealism we created a Blender scene with 8 users - 4 with heavy occlusion, and 4 with less occlusion overlapping the former. In Fig.~\ref{fig6:smart_aggregation} we compare our novel learned federation approach with ablation using FedAvg aggregation strategy trained on this scene. We also compare with FedNeRF. Our learned federation learns to weigh users with less occlusion, boosting PSNR on global content (GT). In real-world scenarios, DecentNeRF would access clients with varying occlusion levels and learn to weigh them explicitly, without predefined heuristics. This allows for better optimization than existing FedAvg.



\subsection{Photorealism and Decentralization analysis}
\noindent\textbf{Novel Blender Dataset.} We do a quantitative and qualitative analysis of photorealism and decentralization on our novel Blender dataset in Table~\ref{tab:blender_baseline} and in Fig \ref{fig:blender_qualitative_utility}. Our results highlight that our approach requires $\sim10^5\times$ server compute than a centralized approach with only a minimal decrease in photorealism as opposed to FedNeRF which breaks in the presence of occlusion/personal content.

\noindent\textbf{Phototourism Dataset.} We do a quantitative comparison of DecentNeRF reconstruction performance on real-world crowdsourced Phototourism Dataset for FedNerF, NeRFW, and our approach. In Fig.~\ref{fig:photo_qualitative}  and Table~\ref{tab:photo_qual} the results tell a compelling story that we are able to perform as well as the best performing centralized approach (NeRFW) in real-world scenes while several orders of magnitude less server compute.

\subsection{Analysis of Personal Content Reconstruction by the Server}

\begin{figure}[t!]
    \centering
    \includegraphics[width=\linewidth]{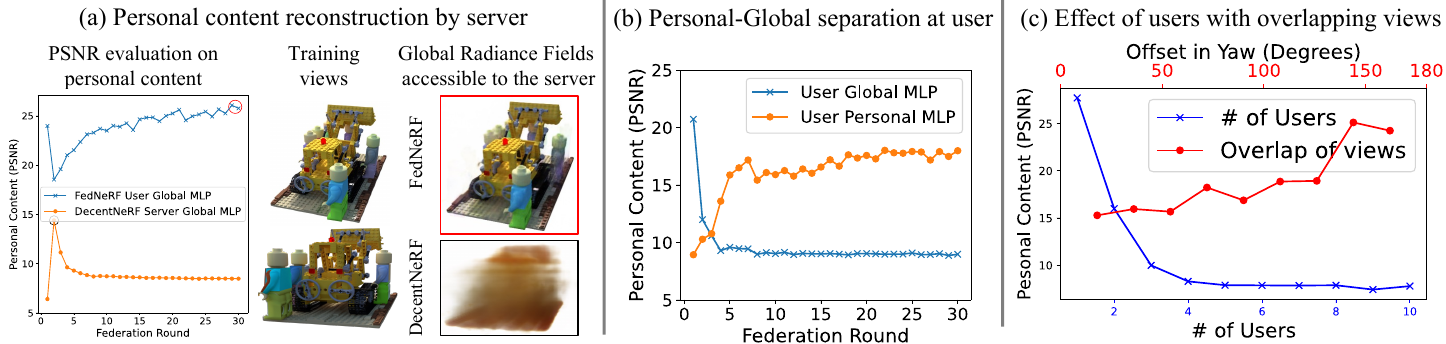}
    \caption{\textbf{Personal content reconstruction}: (a) PSNR evaluation on personal content for server-accessible Radiance fields for FedNeRF and DecentNeRF per round, rendering the worst performance rounds for each on the right. We notice in the server renderings that FedNeRF cannot prevent the server from reconstructing the personal content. At the same time, DecentNeRF reduces the reconstruction of personal content(by 10 dB) even in initial rounds. (b) PSNR evaluation of personal content for users' global and personal MLP for each round demonstrates DecentNeRFs ability to separate personal and global content over multiple rounds of federation. (c) We evaluate the effect of (1) the number of users and (2) the overlap in the views of two users views on the faithful rendering of personal content by the server for DecentNeRF.} 
    \label{fig7:blender_privacy}
\end{figure}

\noindent\textbf{Personal content reconstruction on Blender.} PSNR curves for server-accessible MLPs, i.e., user global MLP for FedNeRF and server global MLP for DecentNeRF in Fig \ref{fig7:blender_privacy} (a) demonstrates a significant reduction of reconstruction (by 10dB) of personal content by the server for DecentNeRF. Visual inspection of server-accessible renderings for DecentNeRF shows the inability of the server to reconstruct the personal content faithfully.

\noindent\textbf{Personal-global separation.} The PSNR curves of users' Global and Personal MLPs on personal content rendering, as shown in Fig \ref{fig7:blender_privacy} (b), demonstrate that personal content initially appears in the users' global MLP. However, over multiple rounds of federation, our two MLP approaches shift the personal content onto the users' personal MLPs. We strongly encourage the reader to inspect the supplemental video. It shows free-view renderings of our personal and global MLPs for different rounds and demonstrates our ability to separate content.

\label{sec:num_users_recons_personal_content}
\noindent\textbf{Worst case reconstruction of personal content.} We notice that the users' global MLP contains personal content during initial rounds. However, because the server can only access the securely aggregated MLPs for DecentNeRF, it cannot reconstruct personal content faithfully due to averaging. We further analyze this worst-case reconstruction of personal content in Fig. \ref{fig7:blender_privacy} (c) by (1) \textit{varying the number of users} with the same overlap in views and (2) \textit{changing the overlap of two users} by changing their view in yaw for the blender "lego" scene. Our analysis highlights two extremes: (1) if there are two users with a significant overlap of views, then the server cannot reconstruct the personal content of either user. (2) If two users have no overlapping views, i.e., they capture entirely separate areas of the scene, then the server's ability to reconstruct personal content increases considerably.

\begin{figure}[t!]
    \centering
    \includegraphics[width=\linewidth]{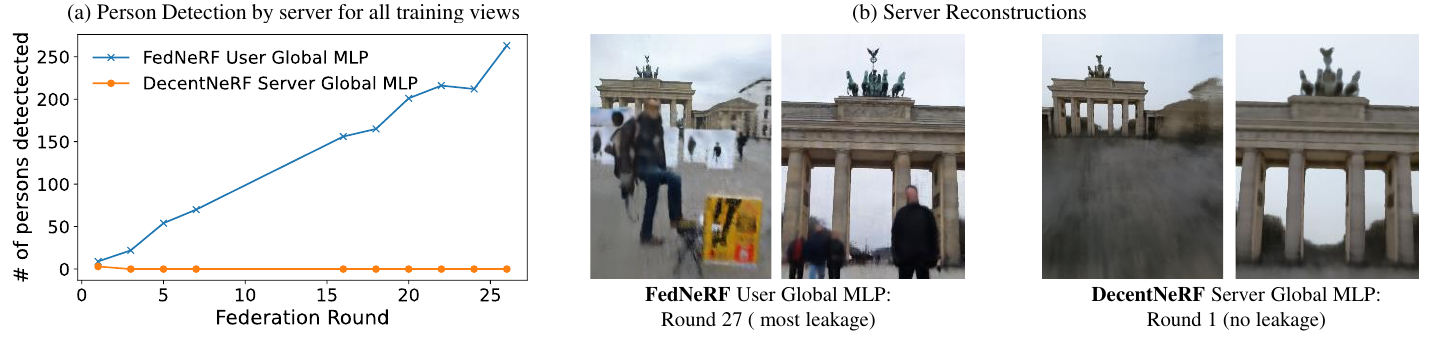}
    \caption{\textbf{Personal content reconstruction on Phototourism Dataset}: (a) Number of persons detected for server-accessible Radiance fields for FedNeRF and DecentNeRF per round for all users (b) Rendering server-accessible Radiance Fields for FedNeRF and DecentNeRF for training views. We note how DecentNeRF reduces the reconstruction of personal content on the server quantitatively and qualitatively.}
    \label{fig5:photo_privacy}
\end{figure}

\noindent\textbf{Personal content reconstruction on Phototourism.} As noted in Fig \ref{fig5:photo_privacy} (a) a person detection NN \cite{girshick2015fast} fails to detect people (personal content) from the server global MLP renderings of DecentNeRF. Corresponding renderings shown in Fig \ref{fig7:blender_privacy} (b) validate that the server renderings do not reconstruct personal content faithfully for DecentNeRF.
\vspace{-10pt}

\section{Discussion}

We present DecentNeRF, a decentralized framework for learning NeRFs, that we believe is crucial for enabling global-scale crowd-sourced NeRFs. DecentNeRF decomposes multi-view images into personal and global visual content. It then securely aggregates the global content across users with high visual fidelity while minimizing the reconstruction of personal content on the server.

\noindent While we analyze DecentNeRF on phototourism scenes, it would enable decentralized NeRF approaches for capturing large-scale 3D scenes and live events, facilitating widespread adoption of NeRFs. DecentNeRF's photorealism and decentralization advantages can be further enhanced by leveraging future advancements in NeRF architectures, training, and rendering schemes. DecentNeRF would pave the way for cross-collaborations between neural rendering and decentralized learning communities.

\noindent\textbf{Limitations.} 
While we do not demonstrate DecentNeRF on actual user mobile devices, there are recent advancements in mobile NeRF renderings \cite{chen2022mobilenerf} that indicate the promise for eventual mobile deployment. DecentNeRF assumes the server can't access the individual user radiance fields from the securely aggregated radiance fields. However, there can be attacks on SMPC \cite{bagdasaryan2020backdoor}  that attempt to break this protection, for which defenses \cite{burkhart2010sepia} have been explored in differential privacy and federated learning literature. Please refer to the supplementary material
on details of these attacks. Incorporating privacy guarantees and more efficient secure decentralization strategies would be exciting future directions.

\clearpage
 

%
%
\bibliographystyle{splncs04}
\bibliography{main}


\end{document}